\documentclass[10pt,twocolumn,letterpaper]{article}

\usepackage[pagenumbers]{cvpr}

\definecolor{cvprblue}{rgb}{0.21,0.49,0.74}
\usepackage[breaklinks,colorlinks,allcolors=cvprblue]{hyperref}
\usepackage{multirow}

\def\confName{CVPR}
\def\confYear{2026}

\title{ZODS-RS — Zero-training Oriented Detection \& Segmentation for Remote Sensing
}

\author{
Zuan Gu \qquad Tianhan Gao\textsuperscript{*} \qquad Langxu Zhao\\
Northeastern University, China\\
{\tt\small 2010497@stu.neu.edu.cn, gth@swc.neu.edu.cn, 2471410@stu.neu.edu.cn}
}
\begin{document}
\maketitle
 \begin{abstract}
Remote-sensing and UAV applications need models that generalize across platforms and viewpoints without task-specific training. Yet training-free pipelines often falter on oriented geometry, scale/rotation variation, and crowded ports or airfields, and rarely unify detection and segmentation.
We introduce ZODS-RS, a training-free, closed-form pipeline that outputs horizontal boxes (HBB) and instance masks. Built on DINOv3 dense features and SAM-style proposals, ZODS-RS chains: PP (prototype purification via Tyler covariance), R-SEM (rotation--scale equivariant matching with separable kernels and global Hungarian assignment), and UAM (uncertainty-aware pixelwise merging with adaptive priors and optional negative prototypes). A lightweight CWLA fuses multiple DINOv3 layers.

On FAIR1M (HBB) we obtain $\mathrm{mAP}_{0.50:0.95}=\mathbf{13.06}$ and $\mathrm{AP}_S=\mathbf{2.93}$ \emph{(class-averaged over ship/airplane)}; on xView (HBB) we report $\mathrm{mAP}=\mathbf{16.69}$. On our UAV dataset, ZODS-RS achieves mask $\mathrm{mIoU}=\mathbf{31.10}$ and improves small-object AP by $\mathbf{+30.70}$ over Grounded-SAM on a single 5090.

This work offers a unified, \emph{no-training} solution for horizontal-box detection plus instance segmentation in aerial imagery; provides explicit closed-form formulations for PP/R-SEM/UAM tightly coupled with DINOv3; and demonstrates \emph{consistent} gains on small and crowded targets and under cross-domain shifts while keeping deployment simple.


\end{abstract}
 \section{Introduction}
\label{sec:intro}

Remote-sensing imagery from satellites and low-altitude UAVs features ultra-high resolutions, pronounced oriented geometry (OBB), heavy scale variation, and crowded layouts across platforms and view angles. These factors make proposal generation and matching difficult, especially when images span thousands of pixels per side and instances are dense or elongated (e.g., ships, aircraft). Public benchmarks codify these challenges: FAIR1M supplies arbitrarily oriented boxes in large tiles, and xView stresses fine-grained categories and small objects under varying resolutions \cite{xview_2018,fair1m_2021}.

The cost of per-task training and cross-domain adaptation motivates \emph{training-free} pipelines, but they often degrade when dense features are not separable, proposal quality is unstable, or cross-image memory is absent. \emph{Open-vocabulary detection} and assembled recipes help via language-aligned encoders or modular segmentors, yet performance remains highly sensitive to backbone–proposal synergy and domain shift \cite{owlv2_2023,grounded_sam_2024,clip_2021,detic_2022,vild_2021,regionclip_2021,ro_vit_2023,glip_2022}. Foundation segmentors like SAM2 provide promptable, high-recall masks widely used as candidates, but their utility hinges on how dense features and downstream matching/merging are organized without training \cite{sam2_2024,sam_2023,samrs_2023,sam_space_2024,sam2_2024}. Persisting gaps are: (i) rotation/scale handling is usually approximate—supervised oriented detectors address it but require (re)training \cite{roi_transformer_2019}; (ii) prototypes are fragile, drifting across scenes without purification or stable sub-prototypes; and (iii) merging is coarse—set-level assignment (e.g., Hungarian) enforces exclusivity but ignores pixel-wise uncertainty, fostering over-confident unions in crowded regions \cite{kuhn_1955,energy_ood_2020,detr_2020,bayes_uncertainty_2017}.

\begin{figure*}[h]
  \centering
  \includegraphics[width=\textwidth]{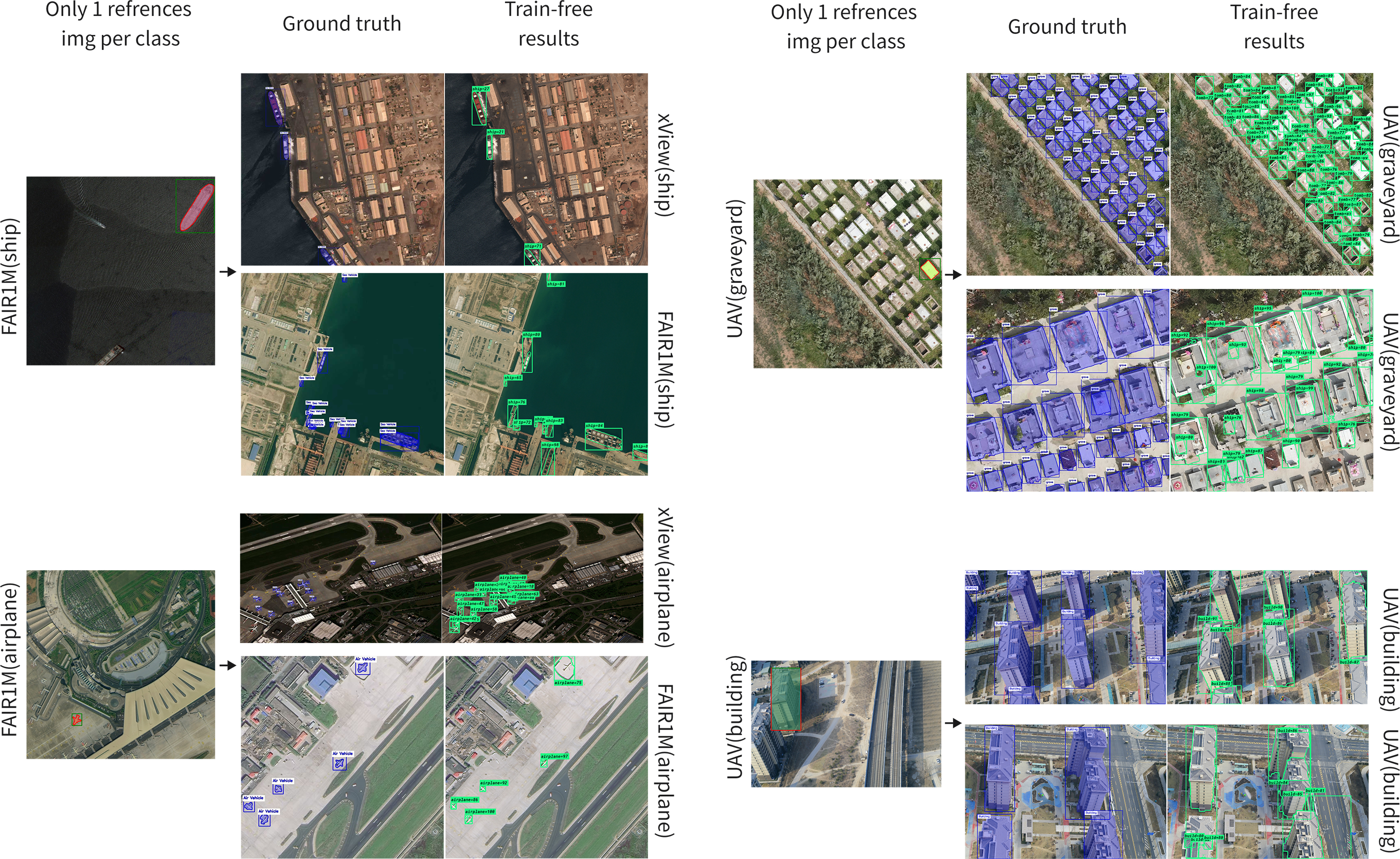}
  \caption{Our method is evaluated directly on different datasets without any fine-tuning\cite{dino_2021,dinov2_2023}. The reference set contains one exemplar image per class. Conditioned on this reference set, the model segments the entire target dataset. The results show: (1) strong generalization across sensing scales—from satellite remote sensing to low-altitude UAV imagery; (2) state-of-the-art performance in 1-shot segmentation without any training or domain adaptation; and (3) effective prediction on images of varying sizes without tiling or upscaling.}
  \label{fig:title}
\end{figure*}

Our view is that frozen-backbone synergy is the foundation for training-free perception in remote sensing: DINOv3 for robust multi-layer dense features, SAM2 for high-recall proposals, and a persistent memory for cross-image/domain retrieval. Building on this backbone triad, ZODS-RS provides a \emph{closed-form, training-free} pipeline that outputs \emph{HBB + instance masks} via three inference-only stages: PP (prototype purification), R-SEM (rotation–scale-equivariant matching with global assignment), and UAM (uncertainty-aware pixel-level merging). We stabilize multi-layer features using a lightweight CWLA,  as a semantic gate to suppress background look-alikes without fine-tuning \cite{dinov3_2025,sam2_2024}. This  work provides a training-free, closed-form DINOv3+SAM2+memory pipeline with a \( \text{PP}\!\rightarrow\!\text{R-SEM}\!\rightarrow\!\text{UAM} \) chain (plus CWLA) producing HBB and masks.

On xView\cite{xview_2018}, FAIR1M\cite{fair1m_2021}, and our custom UAV dataset, ZODS-RS attains HBB mAP\(_{.50:.95}\) of 16.69 (xView), 13.06 (FAIR1M), and 47.30 (UAV). Under a unified HBB protocol, ZODS-RS consistently exceeds training-free/OVD baselines on xView and FAIR1M and yields larger gains on the domain-specific UAV set—indicating practical generalization to bespoke data without any training or adaptation. Our numerics on xView/FAIR1M, mAP\(_{.50:.95}\) improves by +1.37 to +4.47 over the strongest baseline; on the UAV set, by +41.69 \cite{owlv2_2023,grounded_sam_2024}.

Contributions.
\begin{enumerate}
    \item A training-free, closed-form pipeline that unifies detection and instance segmentation (HBB + masks) for aerial/UAV imagery, built on a frozen DINOv3 + SAM2 + memory triad and stabilized by CWLA.
    \item Module-level mathematics tightly coupled to DINOv3: PP with Tyler scatter and spectral purification plus OT-based anchoring; R-SEM with separable rotation/scale weights and Hungarian assignment; UAM with energy-based uncertainty and adaptive priors.
    \item Empirical gains on xView, FAIR1M, and a custom UAV dataset under a unified HBB protocol, with clear benefits on small objects, crowded scenes, and cross-domain robustness—while keeping deployment simple and training-free.
\end{enumerate}
ZODS-RS treats frozen-backbone synergy with memory as a first-class design principle and realizes a \( \text{PP}\!\rightarrow\!\text{R-SEM}\!\rightarrow\!\text{UAM} \) inference chain on top of DINOv3+SAM2+memory, addressing rotation, scale variation, crowding, and domain shifts without fine-tuning.








 \section{Related Work}
\label{sec:related work}

\subsection{Training-free / OVD / Prompted detection}

Recent progress on training-free and open-vocabulary detection (OVD) typically relies on frozen foundation features and strong proposal generators, then aggregates or aligns scores with textual priors to produce pseudo labels or final detections. OWL-V2 scales OVD with web-scale self-training and a text-conditioned detector, showing that robust language–vision priors can substantially reduce reliance on closed vocabularies \cite{owlv2_2023}. Florence-2 offers a unified, prompt-based representation supporting grounding and detection without task-specific fine-tuning \cite{florence2_2024}. In practice, Grounded-SAM composes Grounding DINO with SAM to couple open-set localization and high-quality masks under text prompts, illustrating a widely used two-stage recipe \cite{grounded_sam_2024}. More recently, “No-Time-To-Train” pushes training-free reference-driven segmentation/detection with a staged pipeline anchored by memory and cross-image correspondences \cite{nttt_2025}. Relative to these lines, we treat the frozen backbone synergy—DINOv3 for dense multi-layer features, SAM2 for high-recall proposals, and a persistent memory bank for cross-image/domain retrieval—as a first-class design choice, and address fine-grained/scale/crowding issues with a closed-form PP→R-SEM→UAM chain , rather than viewing the backbone as a replaceable component\cite{clip_2021,detic_2022,vild_2021,regionclip_2021,ro_vit_2023,glip_2022}.

\subsection{Dense Features from DINOv2 to DINOv3}
The DINO family demonstrates that large-scale self-supervised ViTs can provide strong dense features with good transfer across pixel- and region-level tasks. DINOv2 established a robust recipe based on curated data and stabilized training, yielding versatile features for downstream detection/segmentation \cite{dinov2_2023}. DINOv3 further emphasizes high-quality dense alignment and post-hoc flexibility (e.g., resolution/model-size scaling), reporting stronger dense descriptors than prior self-/weakly supervised backbones \cite{dinov3_2025}. In ZODS-RS, we do not merely “swap” backbones; instead, we pair DINOv3 with lightweight cross-layer weighted aggregation (CWLA) and a retrieval memory to enhance separability and detectability under cross-domain shifts, keeping the backbone frozen\cite{dino_2021}.

\subsection{SAM2-based Remote Sensing Generation}
Promptable segmentation models are widely adopted as proposal generators. SAM provides class-agnostic, high-quality masks from sparse prompts \cite{sam_2023,sam_space_2024}, while SAM2 improves temporal stability and streaming efficiency, and is increasingly used as a reusable segmentation engine \cite{sam2_2024}. In aerial/remote-sensing imagery, however, small objects, weak textures, and extreme scales challenge proposal quality; empirical studies on “SAM in RS” highlight mixed results and the need for domain-aware prompting or filtering \cite{samrs_2023}. Our design embeds SAM2 into the backbone synergy: DINOv3’s dense similarity maps guide proposal selection, and R-SEM’s rotation–scale kernels choose candidates that best match purified prototypes

\subsection{Memory Bank \& Prototype Retrieval}
Prototype- or memory-based retrieval is a recurring tool in training-free and weakly supervised settings, but single-mean prototypes and cosine scoring are sensitive to domain shifts and reference selection variance. OVD systems (e.g., OWL-V2) and unified models (e.g., Florence-2) implicitly attest to the utility of semantic priors and external knowledge \cite{owlv2_2023,florence2_2024}. ZODS-RS constructs a reference subset with purified sub-prototypes from stable dense features, using PP to reduce cross-source noise before R-SEM matching\cite{dota_2018,fair1m_2021}\cite{protonet_2017,hdbscan_2017,tyler_1987,cuturi_2013}.

\subsection{Equivariant Matching in Crowded Scene}
Remote-sensing objects exhibit large variations in orientation and scale; classical detectors rely on multi-scale pyramids or incorporate rotation handling explicitly. Rotation-equivariant detection (e.g., ReDet\cite{redet_2021}) integrates equivariant backbones and rotation-invariant pooling for OBB prediction \cite{redet_2021}, while oriented detectors such as RoI Transformer address geometry explicitly at the proposal/ROI level \cite{roi_transformer_2019}. For assignment, the DETR family popularized global bipartite matching with the Hungarian algorithm, providing a clean paradigm for sparse set prediction under overlaps and duplicates \cite{detr_2020}. We follow the matching spirit—using global assignment inside R-SEM—but remain training-free and couple it with closed-form scale–rotation kernels; datasets such as DOTAv2 and FAIR1M highlight the need for such handling in real imagery.

\subsection{Pixel Uncertainty and Bayesian Merging}
Uncertainty modeling distinguishes aleatoric from epistemic components and has improved robustness for dense prediction; Kendall \& Gal formalized per-pixel formulations widely adopted in segmentation \cite{bayes_uncertainty_2017}. For boundary refinement, fully connected Dense-CRF remains an effective, lightweight postprocessor compatible with class-agnostic masks \cite{densecrf_2011}. Our UAM stage uses uncertainty-aware pixel-wise merging with Bayesian weighting and optional CRF refinement to stabilize overlaps and suppress open-set distractors in a training-free regime\cite{energy_ood_2020}.

 \section{Method}
\label{sec:method}

\begin{figure*}[h]
  \centering
  \includegraphics[width=\linewidth]{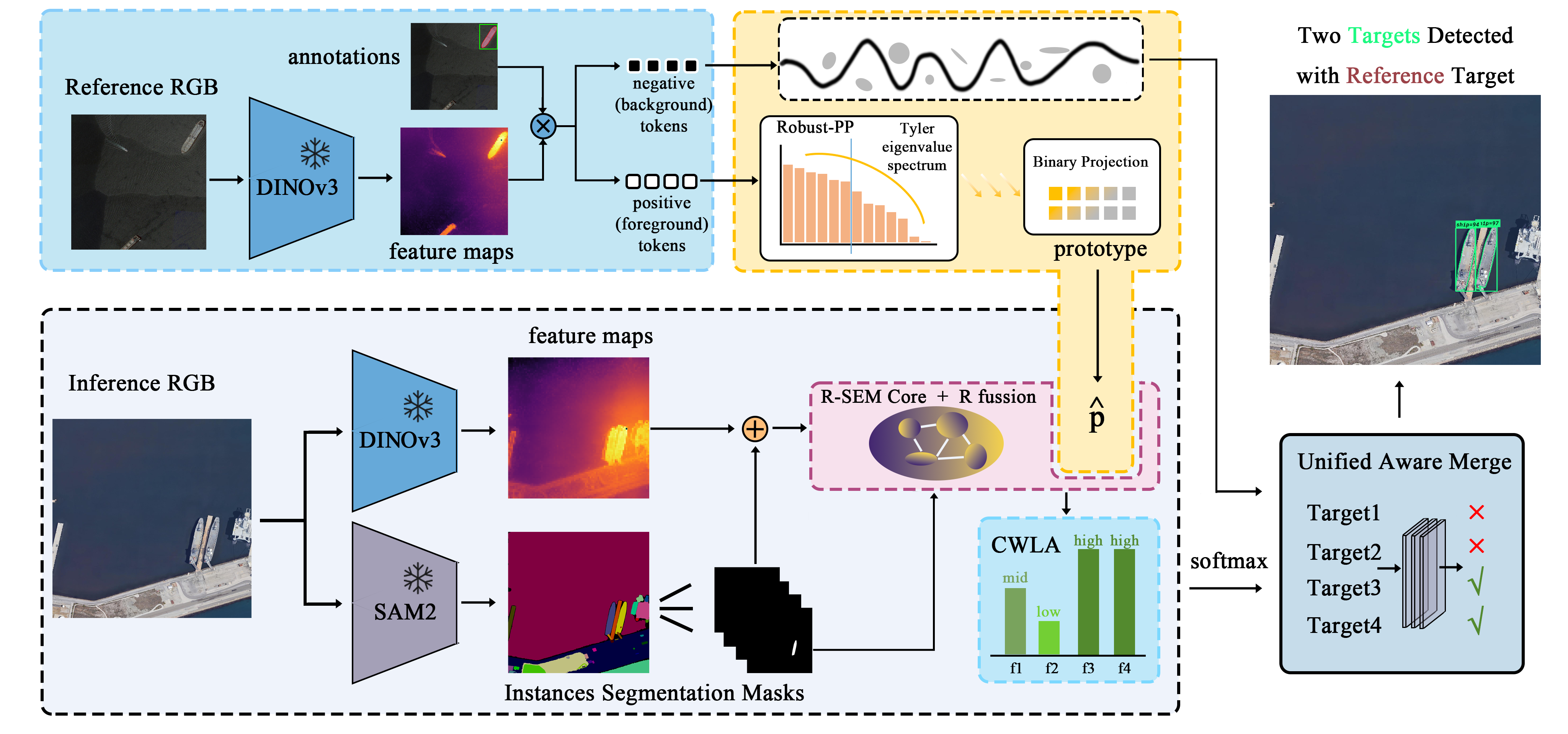}
  \caption{\textbf{Overview of the ZODS-RS pipeline (training-free).}
  \emph{Reference branch} (top-left): DINOv3 extracts dense tokens from a reference RGB patch; annotations split foreground/background tokens.
  Robust-PP estimates a Tyler scatter, applies spectral purification with a binary projection, and yields a purified prototype $\hat{\mathbf p}$ while mining negatives.
  \emph{Inference branch} (bottom): on a test image, DINOv3 provides multi-layer features and SAM2 supplies high-recall instance masks/HBBs.
  R-SEM computes rotation--scale equivariant responses and fuses them; CWLA weights layers by consistency.
  \emph{Merge} (right): UAM performs uncertainty-aware voting over overlapping candidates and suppresses look-alikes with negative channels, producing final HBBs and masks.
  All steps are closed-form or finite-iteration without training.}
  \label{fig:pipeline}
\end{figure*}

\subsection{Engine: DINOv3 + SAM2 + Memory}
\label{subsec:engine}

We adopt a training-free, inference-time engine (show in fig.\ref{fig:pipeline}) that produces geometry-faithful dense features, high-recall proposals, and a persistent memory of purified prototypes and informative negatives, so that \textbf{PP / R-SEM / UAM} operate in a stable, high-SNR space without gradients or EMA.

\textbf{dense tokens.}
Given an image \(I\in\mathbb{R}^{H_0\times W_0\times 3}\), we select \(L\) DINOv3 layers \(\mathcal L=\{\ell_1,\dots,\ell_L\}\).
After resizing to a common grid \((H,W)\) and L2-normalization, the feature tensors and per-pixel tokens are
\begin{equation}
F^{[\ell]}\in\mathbb{R}^{C\times H\times W},\qquad
\mathbf f_p^{[\ell]}\in\mathbb{R}^{C},\qquad
\mathbf f_p^{[\ell]}\leftarrow \mathbf f_p^{[\ell]}/\|\mathbf f_p^{[\ell]}\|_2 .
\end{equation}
A lightweight cross-layer fusion (CWLA; sec.\ref{subsec:CWLA}) later forms \(\bar{\mathbf f}_p\) while preserving layer-wise access for R-SEM.

\textbf{high-recall proposals.}
SAM2 produces masks \(\{M_i\}_{i=1}^{P}\); each mask induces an bounding box (HBB) \(B_i\) via \(\mathrm{minAreaRect}(M_i)\).
We retain masks passing stability checks (minimum area/score, morphology IoU) and perform oriented NMS with IoU threshold \(\tau_{\text{nms}}\), obtaining the set
\begin{equation}
\mathcal M=\{(M_i,B_i)\}_{i=1}^{P'} .
\end{equation}

\textbf{reference-to-memory (R2M).}
A user-marked reference mask—or a self-bootstrapped prediction with high confidence and low uncertainty—is committed into the memory of class \(k\) through a closed-form / finite-iteration routine.
We first collect reference tokens from \(\{F^{[\ell]}\}\) inside the mask and optionally keep only the top-\(q\%\) by cosine similarity (w.r.t. an initial anchor) while eroding borders, yielding a cleaned set \(\mathcal X_k=\{\mathbf x_j\}_{j=1}^{n}\subset\mathbb{R}^{C}\).
We then estimate a scale-free scatter using Tyler’s fixed-point with trace normalization,
\begin{equation}
\Sigma_{t+1}=\frac{d}{n}\sum_{j=1}^{n}
\frac{\mathbf x_j\mathbf x_j^\top}{\mathbf x_j^\top \Sigma_t^{-1}\mathbf x_j},
\qquad
\mathrm{tr}(\Sigma_{t+1})=d,\quad \Sigma_0=I,\ d=C,
\end{equation}
and set \(\widehat{\Sigma}_k=\Sigma_T\), falling back to \(\mathrm{diag}(\Sigma_T)\) or \(I+\varepsilon I\) when small-\(n\) or ill-conditioning is detected.
With \(\widehat{\Sigma}_k=U\Lambda U^\top\), we keep the top-\(r\) axes \(U_r\) and define the idempotent projector
\begin{equation}
P_r=U_r\,\mathrm{diag}(\mathbf w)\,U_r^\top,\qquad \mathbf w\in\{0,1\}^r,\qquad P_r^2=P_r .
\end{equation}
References (and any token \(\mathbf z\)) are whitened and projected,
\begin{equation}
\tilde{\mathbf x}_j=P_r\,\widehat{\Sigma}_k^{-1/2}\,\mathbf x_j,\qquad
\widehat{\Sigma}_k^{-1/2}=U\,\Lambda^{-1/2}\,U^\top ,
\end{equation}
followed by a one-shot bounded-influence mean (M-estimator) and normalization to obtain a purified prototype,
\begin{equation}
\tilde{\mathbf p}_k=\frac{\sum_j a_j\,\tilde{\mathbf x}_j}{\sum_j a_j},\quad a_j=\psi(\|\tilde{\mathbf x}_j\|_2),\qquad
\hat{\mathbf p}_k=\frac{P_r\,\tilde{\mathbf p}_k}{\|P_r\,\tilde{\mathbf p}_k\|_2}.
\end{equation}
When text/image anchors \(\{\mathbf g_m\}\) are available, we optionally shrink \(\tilde{\mathbf p}_k\) toward the anchor manifold via entropy-regularized optimal transport
\begin{equation}
\Pi^\text{*}=\arg\min_{\Pi\in\mathcal U}\langle \Pi, C\rangle+\varepsilon H(\Pi),
\qquad C_{jm}=1-\langle \tilde{\mathbf x}_j,\mathbf g_m\rangle,
\end{equation}
and mix the transport barycenter with weight \(\alpha\) before the final projection/normalization.
To suppress look-alikes, we also mine negatives from low-prior regions (low norm/margin/attention) and aggregate them with the same whiten–project routine, yielding \(\{\tilde{\mathbf p}^{-}_{k,j}\}\).
The committed memory entry is
\begin{equation}
\mathcal B_k=\big(\hat{\mathbf p}_k,\ \widehat{\Sigma}_k,\ \{\tilde{\mathbf p}^{-}_{k,j}\}_{j=1}^{m_k}\big),
\end{equation}
maintained with bounded capacity and closed-form smoothing (count-weighted convex merge), or branched into sub-prototypes when distributions diverge—still training-free.

\textbf{interface.}
The engine outputs \(\mathcal F=\{\bar{\mathbf f}_p\}\), \(\mathcal M=\{(M_i,B_i)\}\), and \(\mathcal B=\{\mathcal B_k\}_k\).
R-SEM consumes \(\hat{\mathbf p}_k\) and per-layer tokens for rotation–scale matching (sec.\ref{subsec:R-SEM});

\subsection{Robust-PP}
\label{subsec:PP}

Given the memory tuple \(\mathcal B_k=(\hat{\mathbf p}_k,\widehat{\Sigma}_k,\{\tilde{\mathbf p}^{-}_{k,j}\})\) and class references \(\mathcal X_k=\{\mathbf x_j\}_{j=1}^{n}\subset\mathbb{R}^C\) from sec.\ref{subsec:engine}, we refine the prototype and scatter in a training-free manner.
We first re-estimate a scale-free scatter using Tyler’s fixed-point with trace normalization:
\begin{equation}
\Sigma_{t+1}=\frac{d}{n}\sum_{j=1}^{n}
\frac{\mathbf x_j\mathbf x_j^\top}{\mathbf x_j^\top\Sigma_t^{-1}\mathbf x_j},
\qquad
\mathrm{tr}(\Sigma_{t+1})=d,\quad \Sigma_0=I,
\label{eq:tyler_pp}
\end{equation}
and set \(\widehat{\Sigma}_k\leftarrow \Sigma_T\) after \(T\) iterations (finite and gradient-free).
For small-\(n\) or ill-conditioned cases, we fall back to \(\widehat{\Sigma}_k\leftarrow \mathrm{diag}(\Sigma_T)\) or \(I+\varepsilon I\) with \(\varepsilon>0\).
Let \(\widehat{\Sigma}_k=U\Lambda U^\top\) be the eigendecomposition with \(\lambda_1\ge\cdots\ge\lambda_C>0\).
We keep top-\(r\) axes \(U_r=[\mathbf u_1,\dots,\mathbf u_r]\) and define an entropy-gated idempotent projector
\begin{equation}
P_r \;=\; U_r\,\mathrm{diag}(\mathbf w)\,U_r^\top,
\qquad \mathbf w\in\{0,1\}^r,
\qquad P_r^2=P_r.
\label{eq:projector_pp}
\end{equation}
All tokens (references or candidates) are \emph{whitened-and-projected}:
\begin{equation}
\tilde{\mathbf x}_j
\;=\;
P_r\,\widehat{\Sigma}_k^{-1/2}\,\mathbf x_j,
\qquad
\widehat{\Sigma}_k^{-1/2}=U\,\Lambda^{-1/2}\,U^\top.
\label{eq:whiten_pp}
\end{equation}
A one-shot bounded-influence mean (M-estimator) yields
\begin{equation}
\tilde{\mathbf p}_k
=
\frac{\sum_{j} a_j\,\tilde{\mathbf x}_j}{\sum_{j} a_j},
\quad 
a_j=\psi\!\big(\|\tilde{\mathbf x}_j\|_2\big),
\qquad
\hat{\mathbf p}_k \leftarrow
\frac{P_r\,\tilde{\mathbf p}_k}{\|P_r\,\tilde{\mathbf p}_k\|_2}.
\label{eq:proto_pp}
\end{equation}

\paragraph{OT-PP (optional).}
When semantics(eg. Clip/Florence)  anchors \(\{\mathbf g_m\}_{m=1}^{M}\subset\mathbb{R}^C\) (text prompts or image anchors) are available, we shrink \(\tilde{\mathbf p}_k\) toward the anchor manifold via entropy-regularized optimal transport:
\begin{equation}
\Pi^\text{*}=\arg\min_{\Pi\in\mathcal U(\mathbf u,\mathbf v)}
\langle \Pi, C\rangle+\varepsilon H(\Pi),
\qquad
C_{jm}=1-\langle \tilde{\mathbf x}_j,\mathbf g_m\rangle.
\label{eq:ot_pp}
\end{equation}
The transport-weighted barycenter
\begin{equation}
\tilde{\mathbf p}_k^{\text{OT}}
=
\frac{\sum_j (\Pi^\text{*}\mathbf 1)_j \,\tilde{\mathbf x}_j}{\sum_j (\Pi^\text{*}\mathbf 1)_j}
\end{equation}
is optionally mixed with weight \(\alpha\) before the final projection/normalization.
\emph{Properties:} the scatter is scale-free; \(P_r\) is idempotent; OT-PP yields anchor-aware shrinkage; all updates are closed-form or finite-iteration without gradients.
\emph{Output:} updated \((\hat{\mathbf p}_k,\widehat{\Sigma}_k)\) (with or without OT anchoring), consumed by R-SEM and UAM.

\subsection{R-SEM}
\label{subsec:R-SEM}

At inference, the detector outputs axis-aligned boxes $C_i$ (HBBs).
For each $C_i$, we use the box as a SAM2 prompt to produce candidate masks $\{M_{i,t}\}$, select a \emph{stable} mask $M_i$ (e.g., highest mask score or largest IoU with the HBB), and derive an oriented box $B_i := \mathrm{minAreaRect}(M_i)$.
We apply minimum area/score and morphology-IoU sanity checks and perform oriented NMS at IoU threshold $\tau_{\text{nms}}$, forming $\mathcal M=\{(M_i,B_i)\}$.
This conversion is training-free and purely geometric.

Let $\bar{\mathbf f}_p$ denote the per-pixel descriptor (Sec.~\ref{subsec:engine}; CWLA in Sec.~\ref{subsec:CWLA}).
We evaluate class-conditioned responses at multiple scales $s\in\mathcal S$ and angles $\theta\in\Theta$:
\begin{equation}
S_k^{(s,\theta)}(p)=\big\langle \bar{\mathbf f}_{p,(s,\theta)},\,\hat{\mathbf p}_k\big\rangle,
\end{equation}
where $\bar{\mathbf f}_{p,(s,\theta)}$ is obtained by scale pyramids and steerable rotations (implemented by resampling/FFT rotation without training).
We apply separable Gaussian weights
\begin{equation}
\begin{aligned}
\alpha^{(s)}&\propto \exp\!\Big(-\tfrac{(\mu^{(s)}-\mu^\ast)^2}{2\gamma^2}\Big),\quad
\beta^{(\theta)}\propto \exp\!\Big(-\tfrac{(\nu^{(\theta)}-\nu^\ast)^2}{2\eta^2}\Big),\\
\bar w^{(s,\theta)}&=\frac{\alpha^{(s)}\beta^{(\theta)}}{\sum_{u,v}\alpha^{(u)}\beta^{(v)}}.
\end{aligned}
\end{equation}
and fuse to a rotation--scale equivariant map
\begin{equation}
R_k(p)=\sum_{s\in\mathcal S}\sum_{\theta\in\Theta}\bar w^{(s,\theta)}\,\mathrm{Up}\!\big(S_k^{(s,\theta)}\big)(p).
\end{equation}
For each $(M_i,B_i)\in\mathcal M$, we define a cost balancing saliency, coverage, shape, and orientation:
\begin{equation}
\begin{aligned}
\mathcal C_{i,k}
&= -\,\mathrm{TopKMean}_{K}\!\big(R_k\!\mid\!M_i\big)
-\lambda_{\text{cov}}\,\mathrm{Cover}\!\big(R_k\!\mid\!M_i, t_{\text{cov}}\big)\\
&\quad+\lambda_{\text{shape}}\,\Phi(B_i)
+\lambda_\theta\,\Delta_\theta(B_i;R_k).
\end{aligned}
\end{equation}
Here $\mathrm{TopKMean}_{K}(R\!\mid\!M)$ averages the top-$K$ responses inside $M$,
$\mathrm{Cover}(R\!\mid\!M,t)=|\{p\in M: R(p)\ge t\}|/|M|$,
$\Phi(B_i)$ encodes aspect-ratio/compactness priors,
and $\Delta_\theta$ penalizes the deviation between the OBB angle and the dominant response angle
$\theta^\ast(k)=\arg\max_{\theta}\sum_{p\in M_i} S_k^{(s^\ast,\theta)}(p)$.
We form a bipartite cost matrix $[\mathcal C_{i,k}]$ and solve a one-to-one assignment using the Hungarian algorithm (per image or tile).

\subsection{CWLA (Consistency-Weighted Layer Aggregation)}
\label{subsec:CWLA}

Given per-layer responses \(R_k^{[\ell]}(p)\) computed from tokens \(F^{[\ell]}\) and prototype \(\hat{\mathbf p}_k\), we define a layer-level uncertainty \(\overline{U}^{[\ell]}\) (e.g., spatially-averaged entropy of a temperature-scaled softmax over \(k\), or energy \(-\tau\log\sum_{k}\exp(R_k^{[\ell]}/\tau)\)).
We convert uncertainties into soft weights
\begin{equation}
\beta_\ell=\mathrm{softmax}\!\left(-\frac{\overline{U}^{[\ell]}}{\sigma}\right),
\qquad \sum_{\ell\in\mathcal L}\beta_\ell=1,\ \beta_\ell\ge 0,
\label{eq:cwla_weights}
\end{equation}
and obtain the fused response
\begin{equation}
R_k(p)=\sum_{\ell\in\mathcal L}\beta_\ell\ \mathrm{Up}\!\big(R_k^{[\ell]}\big)(p).
\label{eq:cwla_fuse}
\end{equation}
This aggregation is closed-form, training-free, and interfaces R-SEM by replacing \(R_k^{[\ell]}\) with the equivariant responses prior to fusion.
The temperature \(\sigma\) controls selectivity with respect to layer reliability.

\subsection{UAM }
\label{subsec:UAM}

Let \(R_c(p)\) be the fused response of channel \(c\) (class prototypes plus \emph{negative} channels; negatives come from \(\{\tilde{\mathbf p}^{-}_{k,j}\}\) and are never emitted).
We build pixel logits with adaptive priors \(A_c(p)>0\):
\begin{equation}
L_c(p)=\frac{R_c(p)}{\tau}+\gamma\log A_c(p),
\qquad
\pi_c(p)=\frac{\exp(L_c(p))}{\sum_{q}\exp(L_q(p))}.
\label{eq:uam_logits}
\end{equation}
Define pixel uncertainty \(U(p)=-\sum_c \pi_c(p)\log\pi_c(p)\).
For overlapping instances \(\{M_j\}\) assigned to classes \(\{k_j\}\) (sec.\ref{subsec:R-SEM}), we merge by uncertainty-aware voting:
\begin{equation}
\label{eq:uam_merge}
\begin{aligned}
w_j(p) &\propto \frac{\pi_{k_j}(p)\,A_{k_j}(p)^\gamma}{1+\lambda\,U(p)},\\
\hat{y}(p) &= \arg\max_{k}\sum_{j:\,k_j=k} w_j(p)\,\mathbb{1}\{p\in M_j\}.
\end{aligned}
\end{equation}
Open-set rejection uses either a maximum posterior threshold \(\max_c\pi_c(p)<\tau_{\text{open}}\) or an energy gate \(-\tau\log\sum_c\exp(R_c(p)/\tau)>\epsilon_{\text{open}}\).
A lightweight CRF can optionally refine boundaries while keeping scores fixed (no training).
Negatives act only as competing channels in the softmax, suppressing look-alikes.

 \section{Experiments}
\label{sec:exp}

\subsection{Experiments Setup}
\subsubsection{Datasets}
We evaluate on xView, FAIR1M, and a self-collected UAV set; none is used for training. To harmonize labels, we merge xView/FAIR1M to {\textit{plane}, \textit{ship}} and UAV to {\textit{graveyard}, \textit{building}}. All annotations are expressed as HBB; when OBBs exist, we convert them to axis-aligned enclosing boxes (AABB). For mask-capable methods, we derive SAM2-prompted proxy masks from HBBs (quality-checked) and use them only for evaluation.

\subsubsection{Task \& Protocol (single-image, no tiling)}
All methods run a single forward pass per image with no cropping/tiling. The primary task is HBB detection. For approaches that also produce masks (e.g., training-free segmentation pipelines), we additionally report mask metrics using the SAM2-derived proxy masks.When open-vocabulary prompts are required, all baselines share the same canonical prompt list per merged class (e.g., ship,airplane,graveyard) and identical text pre-processing.

\subsubsection{Metrics}
 We report COCO-style HBB $\mathbf{mAP_{0.50:0.95}}$ , $\mathrm{AP}_{50}$, $\mathrm{AP}_{75}$, and $\mathrm{AP}_{S}$ (small-object AP). 
For methods that output masks, we additionally report per-instance mask IoU and Dice.

\subsection{Baselines and Implementation Details}
\subsubsection{Baselines}
We compare against No-Time-To-Train (NTTT), Grounded-SAM (Grounding-DINO + SAM), OWL-V2, LAE-DINO, and Florence-2. Unless otherwise stated, we use each method’s official inference settings and evaluate HBB detection under the unified protocol in Sec. 4.1 (single-scale, no tiling, shared NMS/thresholds, identical prompt lists for open-vocabulary variants). For OWL-V2, we tune the score threshold to 0.30 on a small validation subset and keep all other parameters aligned with the shared protocol.
Because LAE-DINO is pre-trained on LAE-1M, which includes FAIR1M (via LAE-FOD) and xView, evaluating on those sets would cause train–test overlap. To keep the protocol disjoint, we therefore report results in this subsection \emph{only} on our self-collected UAV dataset and omit FAIR1M/xView for LAE-DINO comparisons.

\subsubsection{Implementation Details (condensed)}
ZODS-RS follows the engine in sec.\ref{subsec:engine} and the modules in sec.\ref{subsec:PP}-- sec.\ref{subsec:UAM}: frozen DINOv3 dense tokens, SAM2 high-recall proposals, and a persistent memory bank feeding Robust-PP, optional R-SEM, CWLA, and UAM, all in a training-free regime. Concretely, we use DINOv3 (sat-ViT-L) with multi-layer tokens from layers \([8,10,-1]\) and an attention prior; SAM2 is \emph{hiera-large} with \texttt{points\_per\_side=32}, \texttt{iou\_thr=0.4}, \texttt{nms\_thr=0.5}, and \texttt{num\_out\_instance=100}. Robust-PP employs Tyler scatter (\texttt{iters=20}, \texttt{tol=1e-4}), spectral truncation with \texttt{top\_r=32}. CWLA is enabled over layers \([8,12,-1]\) using the \emph{agreement} metric with temperature \(\sigma=0.15\) and auto-temperature active, yielding an uncertainty-aware fusion that remains gradient-free.

UAM uses temperature \(\tau=1.0\) with \emph{auto-\(\tau\)} (entropy target \(0.9\), range \([0.5,2.0]\)), combines priors in the \emph{logit} domain with strength \(\gamma=1.5\) and component weights \((a,b,c)=(0.5,0.5,0.2)\), enables negative prototypes (\texttt{num=3}, \texttt{min\_area=128}) to suppress look-alikes, and applies a lightweight CRF (\texttt{n\_iters=10}, \texttt{sxy=[3,3]}, \texttt{srgb=[5,5,5]}) for boundary refinement.  HBB-NMS IoU \(\tau_{\text{nms}}=0.50\), and mask binarization \(\tau_{\text{mask}}=0.50\). Inference runs on a single 5090 GPU in FP32.

\subsection{Experimental Results}
\label{sec:exp_results}

\begin{figure}[h]
  \centering
  \includegraphics[width=\linewidth]{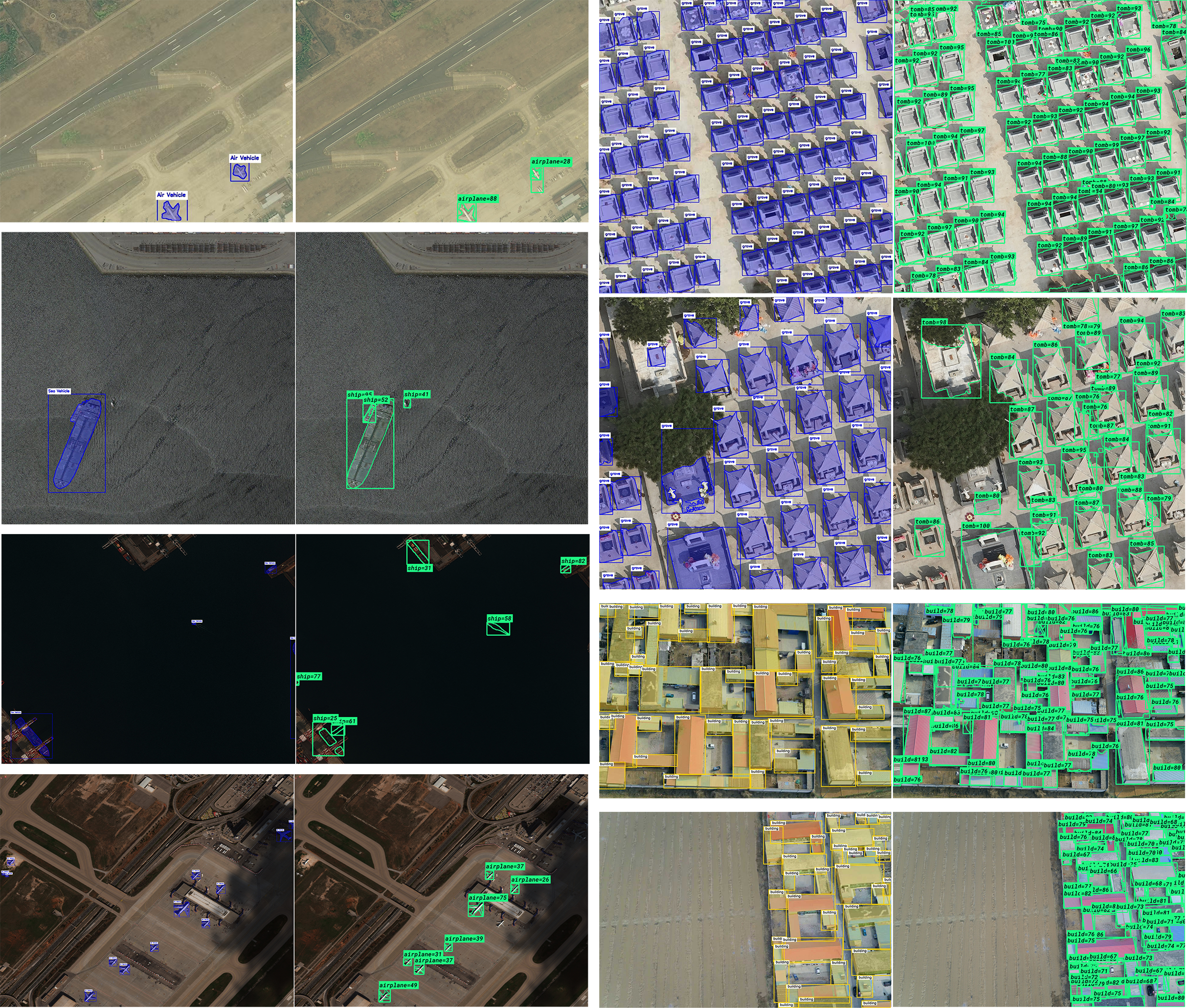}
  \caption{One-shot cross-domain results. With full-image inference, ZODS-RS cleanly separates overlapping instances in crowded scenes while preserving fine-grained localization, enabled by PP, R-SEM, CWLA, and UAM.}
  \label{fig:re1}
\end{figure}

Evaluation setup and notes.
We adopt a \emph{training-free, cross-domain} protocol. For \textit{ship}/\textit{airplane}, one FAIR1M reference per class is used to evaluate on FAIR1M and xView; for UAV (graveyard), the first image serves as the reference and we evaluate the full set. Inference is \emph{single-scale} on full images (no tiling/upscaling). Baselines use default thresholds except OWL-V2 (score = 0.30). We report COCO-style HBB metrics (mAP\(_{0.50:0.95}\), AP50, AP75, APs); IoU/Dice only for mask-capable methods. Because LAE-DINO is pre-trained on a corpus including xView/FAIR1M, it is reported only on UAV to avoid train–test overlap.

UAV Data.
Overall, ZODS-RS achieves mAP 47.30, far surpassing all baselines (next best Florence-2: 5.61, i.e., \(\approx8.4\times\)). AP50 is 50.31 (\(\approx8.3\times\)), AP75 48.25 (\(\approx8.0\times\)), and APs 33.65 (\(\approx11.4\times\)). On masks, ZODS-RS reaches IoU 31.10 vs. Grounded-SAM 16.62 (\(\approx1.87\times\)), and Dice 42.71 vs. 24.24 (\(\approx1.76\times\)). LAE-DINO scores mAP 0.93 and AP50/75 1.27, trailing ZODS-RS by \(>\!50\times\). These results indicate that \emph{1-shot memory + full-image inference} provides strong utility in bespoke, non-textbook distributions (Tab.~\ref{tab:main_results}, Fig.~4).

FAIR1M (reference from FAIR1M). \emph{Ship.} ZODS-RS attains mAP 23.93, outperforming Florence-2 20.38, OWL-V2 16.32, Grounded-SAM 15.55, and NTTT 6.22. Relative to the strongest baseline (Florence-2), gains are +3.55 mAP (+17\%), +13.78 AP50 (+47\%), and +8.01 AP75 (+40\%). On APs, OWL-V2 5.93 exceeds ours 4.84 (\(-1.09\)), suggesting that extremely small targets remain a recall bottleneck under our no-tiling constraint. For masks, ZODS-RS yields IoU 56.57 / Dice 67.92, clearly above Grounded-SAM 33.26 / 41.79 and NTTT 21.21 / 29.46, reflecting stronger boundary fidelity and consistency. \emph{Airplane.} Grounded-SAM reaches the highest mAP 3.01; ours 2.19, Florence-2 2.01, OWL-V2 1.62, NTTT 0.82. AP50 follows the same order (\(16.22 > 11.96\) for ours). Under the stricter AP75, ZODS-RS scores 0.67, exceeding Florence-2 0.28, OWL-V2 0.13, and Grounded-SAM 0.00, indicating more precise localization at high IoU despite lower overall mAP. Mask quality is also superior (IoU 55.15 / Dice 68.79 vs. \(23.76/29.66\) for Grounded-SAM and \(16.72/19.36\) for NTTT). (Tab.~\ref{tab:main_results}) 

xView (reference from FAIR1M; cross-domain). \emph{Ship (cross-domain).} ZODS-RS leads with mAP 13.07, ahead of Florence-2 11.45, Grounded-SAM 8.07, OWL-V2 7.16, NTTT 3.77; AP50 18.43 and AP75 15.10 are also highest. For APs, OWL-V2 2.03 slightly surpasses ours 1.82 (\(-0.21\)). On masks, Grounded-SAM IoU 22.25 \(>\) ours 19.36, while Dice ours 25.02 \(>\) 23.52, hinting that boundary coherence benefits from our merging even when raw overlap lags. \emph{Airplane (cross-domain).} ZODS-RS attains mAP 20.31, substantially above Florence-2 13.00, Grounded-SAM 4.73, OWL-V2 4.07, NTTT 1.97 (\(+56\%\) over Florence-2). AP50 39.90 and AP75 24.62 are likewise dominant (AP75 is \(\approx3.4\times\) the next best). Masks reach IoU 24.03 / Dice 33.62, comfortably exceeding Grounded-SAM 15.96 / 10.03. (Tab.~\ref{tab:main_results})

Summary.
(i) On UAV data, ZODS-RS shows large, consistent gains across detection and segmentation, directly supporting the “user-specific dataset” use case. (ii) In \emph{cross-domain} settings (FAIR1M\(\rightarrow\)xView), ZODS-RS is leading or tied on mAP/AP50/AP75 for both \textit{ship} and \textit{airplane}, with particularly strong margins on xView–airplane. (iii) AP75 and mask quality (IoU/Dice) repeatedly favor ZODS-RS, suggesting accurate localization and stable boundaries—even when overall mAP is not the best (FAIR1M–airplane). (iv) APs lags slightly on \textit{ship} for FAIR1M/xView relative to OWL-V2, consistent with the \emph{no-tiling} evaluation; denser scale grids or lightweight zoom-in heuristics are promising remedies. (v) Under a unified \emph{zero-training} protocol with \emph{1-shot reference}, ZODS-RS delivers reproducible, across-dataset gains.

\begin{table}[t]
\centering
\scriptsize
\setlength{\tabcolsep}{3pt}
\renewcommand{\arraystretch}{0.95}
\caption{HBB results. “ship/airplane/graveyard” indicate the active class (1=active). IoU/Dice only for mask-capable methods. OWL-V2 uses score thr.=0.30. LAE-DINO reported only on UAV.}
\label{tab:main_results}
\resizebox{\linewidth}{!}{
\begin{tabular}{l l c c c r r r r r r}
\toprule
Method & Dataset & ship & airplane & graveyard & mAP & AP50 & AP75 & APs & IoU & Dice \\
\midrule
\multirow{5}{*}{OWL-V2 (0.30)} 
 & UAV (graveyard) & 0 & 0 & 1 & 2.92 & 3.52 & 3.52 & 4.23 & -- & -- \\
 & FAIR1M (ship)   & 1 & 0 & 0 & 16.32 & 29.05 & 15.58 & 5.93 & -- & -- \\
 & FAIR1M (airplane) & 0 & 1 & 0 & 1.62 & 8.00 & 0.13 & 0.13 & -- & -- \\
 & xView (ship)    & 1 & 0 & 0 & 7.16 & 13.34 & 8.75 & 2.03 & -- & -- \\
 & xView (airplane)& 0 & 1 & 0 & 4.07 & 11.74 & 4.55 & 0.00 & -- & -- \\
\midrule
\multirow{5}{*}{NTTT} 
 & UAV (graveyard) & 0 & 0 & 1 & 1.53 & 1.33 & 1.33 & 0.00 & 10.10 & 13.52 \\
 & FAIR1M (ship)   & 1 & 0 & 0 & 6.22 & 12.85 & 0.00 & 0.00 & 21.21 & 29.46 \\
 & FAIR1M (airplane) & 0 & 1 & 0 & 0.82 & 9.25 & 0.00 & 0.00 & 16.72 & 19.36 \\
 & xView (ship)    & 1 & 0 & 0 & 3.77 & 11.01 & 1.24 & 0.00 & 8.96 & 11.25 \\
 & xView (airplane)& 0 & 1 & 0 & 1.97 & 5.21 & 0.00 & 0.00 & 5.96 & 11.48 \\
\midrule
\multirow{5}{*}{Florence-2} 
 & UAV (graveyard) & 0 & 0 & 1 & 5.61 & 6.06 & 6.06 & 0.00 & -- & -- \\
 & FAIR1M (ship)   & 1 & 0 & 0 & 20.38 & 29.55 & 19.70 & 0.00 & -- & -- \\
 & FAIR1M (airplane) & 0 & 1 & 0 & 2.01 & 10.23 & 0.28 & 0.00 & -- & -- \\
 & xView (ship)    & 1 & 0 & 0 & 11.45 & 12.60 & 9.50 & 0.00 & -- & -- \\
 & xView (airplane)& 0 & 1 & 0 & 13.00 & 28.69 & 7.20 & 0.00 & -- & -- \\
\midrule
\multirow{5}{*}{LAE-DINO} 
 & UAV (graveyard) & 0 & 0 & 1 & 0.93 & 1.27 & 1.27 & 0.00 & -- & -- \\
 & FAIR1M (ship)   & 1 & 0 & 0 & -- & -- & -- & -- & -- & -- \\
 & FAIR1M (airplane) & 0 & 1 & 0 & -- & -- & -- & -- & -- & -- \\
 & xView (ship)    & 1 & 0 & 0 & -- & -- & -- & -- & -- & -- \\
 & xView (airplane)& 0 & 1 & 0 & -- & -- & -- & -- & -- & -- \\
\midrule
\multirow{5}{*}{Grounded-SAM} 
 & UAV (graveyard) & 0 & 0 & 1 & 5.28 & 5.87 & 5.87 & 2.95 & 16.62 & 24.24 \\
 & FAIR1M (ship)   & 1 & 0 & 0 & 15.55 & 21.13 & 19.96 & 0.00 & 33.26 & 41.79 \\
 & FAIR1M (airplane) & 0 & 1 & 0 & 3.01 & 16.22 & 0.00 & 0.00 & 23.76 & 29.66 \\
 & xView (ship)    & 1 & 0 & 0 & 8.07 & 14.02 & 9.09 & 0.00 & 22.25 & 23.52 \\
 & xView (airplane)& 0 & 1 & 0 & 4.73 & 11.82 & 0.00 & 0.00 & 15.96 & 10.03 \\
\midrule
\multirow{5}{*}{Ours} 
 & UAV (graveyard) & 0 & 0 & 1 & 47.30 & 50.31 & 48.25 & 33.65 & 31.10 & 42.71 \\
 & FAIR1M (ship)   & 1 & 0 & 0 & 23.93 & 43.33 & 27.97 & 4.84 & 56.57 & 67.92 \\
 & FAIR1M (airplane) & 0 & 1 & 0 & 2.19 & 11.96 & 0.67 & 1.01 & 55.15 & 68.79 \\
 & xView (ship)    & 1 & 0 & 0 & 13.07 & 18.43 & 15.10 & 1.82 & 19.36 & 25.02 \\
 & xView (airplane)& 0 & 1 & 0 & 20.31 & 39.90 & 24.62 & 0.66 & 24.03 & 33.62 \\
\bottomrule
\end{tabular}
} 
\end{table}

\subsection{Ablation Studies}
\label{sec:ablation}

We analyze the stepwise design \textit{base} $\rightarrow$ \textit{pp} $\rightarrow$ \textit{pp+sem-cwla} $\rightarrow$ \textit{all} (\textit{pp+sem-cwla+uam}) under the unified, training-free protocol (single-scale, full-image, 1-shot reference; HBB metrics).As show in Tab.~\ref{tab:ablation_main}, on the user-defined UAV set, performance grows from \textit{base} to \textit{pp} and then jumps markedly at \textit{pp+sem-cwla}, culminating in a large margin at \textit{all(pp+sem-cwla+uam)}. Concretely, \textit{base} yields $(0.74/0.74/0.74)$ for mAP/AP50/AP75, \textit{pp} lifts them to $(2.93/2.97/2.97)$, \textit{pp\_sem+cwla} to $(16.68/17.21/16.86)$, and \textit{all} to $(\mathbf{47.30}/\mathbf{50.31}/\mathbf{48.25})$. This trajectory indicates: (i) \textit{pp} aligns reference and candidate features into a more comparable scatter, improving both relaxed and strict IoUs; (ii) adding \textit{sem+cwla} further stabilizes matching through rotation/scale handling and cross-layer consistency; and (iii) the full system with \textit{uam} consolidates overlapping candidates via uncertainty-aware merging and prior reweighting, pushing detection quality and geometric fidelity jointly.

On FAIR1M and xView (cross-domain), trends are consistent but express different trade-offs between recall and localization. On FAIR1M, \textit{pp} sharply increases AP50 from $0.00$ to $18.18$ while mAP fluctuates, \textit{pp+sem-cwla} stabilizes mAP near the \textit{base} level, and \textit{all} turns on strict localization with a non-zero AP75 ($0.67$) despite a reduction in AP50—suggesting a more conservative recall but improved high-IoU precision.
In contrast, cross-domain xView remains low through \textit{pp} and \textit{pp+sem-cwla}, and only with \textit{all} does it surge to $(20.31/39.90/24.62)$, indicating that the combination of equivariant matching, consistency-weighted fusion, and uncertainty-aware merging is essential when domain shift is pronounced: both recall (AP50) and strict localization (AP75) rise together once the full pipeline is active. Overall, the ablation shows a coherent mechanism: \textit{pp} primarily rectifies distribution mismatch (recall-first), \textit{sem-cwla} enforces geometrically consistent correspondences and robust cross-layer evidence, and \textit{uam} regularizes pixel-level competition and overlap to yield stable instance decisions across domains.

\begin{table}[h]
\centering
\scriptsize
\setlength{\tabcolsep}{3pt}
\renewcommand{\arraystretch}{0.95}
\caption{Ablation results. Stepwise evaluation on three datasets; values are mAP/AP50/AP75.}
\label{tab:ablation_main}
\resizebox{\linewidth}{!}{
\begin{tabular}{c l r r r}
\toprule
Model & Dataset  & mAP & AP50 & AP75 \\
\midrule
$\text{Base}$ & FAIR1M  & 3.30 & 0.00 & 0.00 \\
$\text{Base}$ & xView   & 1.60 & 2.86 & 1.33 \\
$\text{Base}$ & UAV     & 0.74 & 0.74 & 0.74 \\
\midrule
$\text{Base+PP}$ & FAIR1M  & 2.76 & 18.18 & 0.00 \\
$\text{Base+PP}$ & xView   & 1.36 & 1.82 & 1.82 \\
$\text{Base+PP}$ & UAV     & 2.93 & 2.97 & 2.97 \\
\midrule
$\text{Base+PP+SEM-CWLA}$ & FAIR1M  & 3.16 & 19.89 & 0.00 \\
$\text{Base+PP+SEM-CWLA}$ & xView   & 1.62 & 3.07 & 1.24 \\
$\text{Base+PP+SEM-CWLA}$ & UAV     & 16.68 & 17.21 & 16.86 \\
\midrule
$\text{Base+PP+SEM-CWLA+UAM}$ & FAIR1M  & 2.19 & 11.96 & 0.67 \\
$\text{Base+PP+SEM-CWLA+UAM}$ & xView   & 20.31 & 39.90 & 24.62 \\
$\text{Base+PP+SEM-CWLA+UAM}$ & UAV     & 47.30 & 50.31 & 48.25 \\
\bottomrule
\end{tabular}
} 
\end{table}
 \section{Conclusion and Future Work}
In this work, we presented ZODS-RS, a training-free, closed-form pipeline for aerial perception that couples frozen DINOv3 dense tokens, SAM2 proposals, and a persistent memory with three inference-only operators—PP for prototype purification, R-SEM for rotation–scale-equivariant matching with global assignment, and UAM for uncertainty-aware pixelwise merging—stabilized by lightweight CWLA. Under a unified HBB protocol , ZODS-RS yields consistent gains over training-free/OVD baselines: mAP$_{0.50:0.95}$ of 16.69 on xView and 13.06 on FAIR1M, and 47.30 on our UAV set, where mask mIoU reaches 31.10 and small-object AP improves by +30.70 over Grounded-SAM. Ablations (base $\rightarrow$ pp $\rightarrow$ pp\_sem+cwla $\rightarrow$ all) corroborate the intended roles of distribution alignment, equivariant matching with cross-layer consistency, and uncertainty-driven merging.

For future work, we aim to extend HBB outputs to true OBB and shape-aware masks within UAM; improve small-object recall via lightweight tiling/zoom-in and denser R-SEM scale/angle grids; adapt priors and memory online (automatic $\tau,\gamma,\lambda$ calibration; retrieval-aware sub-prototype growth/pruning); tighten semantic gating with Florence/CLIP for open-set distractors; leverage SAM2 streaming and short-term tracklets for temporal stability; further optimize deployment with ONNX/TensorRT, kernel fusion, and mixed precision; broaden evaluation across additional RS/UAV corpora with released reference splits; and strengthen reliability through calibrated uncertainty, abstention, and optional human-in-the-loop review for safety-critical workflows.

\end{document}